\documentclass[a4paper,12pt,fleqn]{cas-sc}

\usepackage[authoryear,longnamesfirst]{natbib}
\usepackage{graphicx}
\usepackage{pifont}
\usepackage{subcaption}
\usepackage{float}
\usepackage{array}
\usepackage{xcolor}
\usepackage{hyperref}
\usepackage{multirow}   
\usepackage{booktabs}   
\usepackage{amsmath,amssymb,amsfonts}
\usepackage{textcomp}

\usepackage{url}
\usepackage{cleveref}
\usepackage{xurl}
\newcommand{\tick}{\ding{51}}
\newcommand{\cross}{\ding{55}}

\usepackage{setspace}
\usepackage{geometry}
\def\tsc#1{\csdef{#1}{\textsc{\lowercase{#1}}\xspace}}
\tsc{WGM}
\tsc{QE}
\tsc{EP}
\tsc{PMS}
\tsc{BEC}
\tsc{DE}

\begin{document}
	\let\WriteBookmarks\relax
	\def\floatpagepagefraction{1}
	\def\textpagefraction{.001}
	\doublespacing

		\shorttitle{Predictive Modelling of Solid Waste Generation through Time Series Analysis}

	\shortauthors{M.T. Dey et~al.}

	\title [mode = title]{Predictive Modelling of Solid Waste Generation through Time Series Analysis}                   
	\author[1,2]{Mitra Tithi Dey}[
	orcid=0009-0002-1373-8787]
	
	\cormark[1]
	
	\ead{mitratithidey@nndc.ac.in}
	
	\affiliation[1]{organization={Dept. of Computer Science, Netaji Nagar Day College},
		city={Kolkata},
		country={India}}
	
	\affiliation[2]{organization={School of Mobile Computing and Communication, Jadavpur University},
		city={Kolkata},
		country={India}}
	
	\author[2]{Punyasha Chatterjee}[%
	orcid=0000-0002-8787-6923]
	\ead{punyasha.chatterjee@jadavpuruniversity.in}

	\author[3]{Amlan Chakrabarti}[%
	orcid=0000-0003-4380-3172]
	\ead{acakcs@caluniv.ac.in}
	
	\affiliation[3]{organization={A.K. Choudhury School of Information Technology, University of Calcutta},
		city={Kolkata},
		country={India}}
	\cortext[cor1]{Corresponding author}
	\begin{abstract}
		Reliable forecasting of Municipal Solid Waste (MSW) production is pivotal for sustainable urban management, as it enables efficient resource planning, optimized collection schedules, and reduced environmental impacts. Existing works on MSW forecasting have some limitations, such as unavailability of granular data from lower administrative units, small datasets, limited or sometimes no consideration of important factors that can affect MSW generation, including holidays, weekends, seasonality, changes in temperature or humidity.
	This work presents a comprehensive comparative analysis of four popularly used time series forecasting models, namely Seasonal Autoregressive Integrated Moving Average with Exogenous Regressors (SARIMAX), Prophet, Long Short Term Memory (LSTM) and Gated Recurrent Unit (GRU), using a real-world MSW dataset from Sri Lanka spanning 2012-2018. A detailed factor-wise analysis is conducted to examine waste generation behaviour considering holidays, weekday-weekend variations, seasonal patterns, temperature and humidity. The results show lower waste generation during holidays and weekends, higher generation during Summer and Winter, and comparatively lower generation during Monsoon and Post-Monsoon. Temperature and humidity do not exhibit a direct linear relationship with waste generation, although their seasonal variations are associated with the changes in waste generation across different seasons. All model performances are evaluated using Mean Absolute Error (MAE), Mean Absolute Percentage Error (MAPE), Root Mean Squared Error (RMSE), and the coefficient of determination ($R^2$). Among the evaluated models, LSTM and SARIMAX achieve the best overall forecasting performance.

	\end{abstract}


	\maketitle
	
	\section{Introduction}
	
\begin{figure}
		\centering
		\includegraphics[width=0.9\columnwidth]{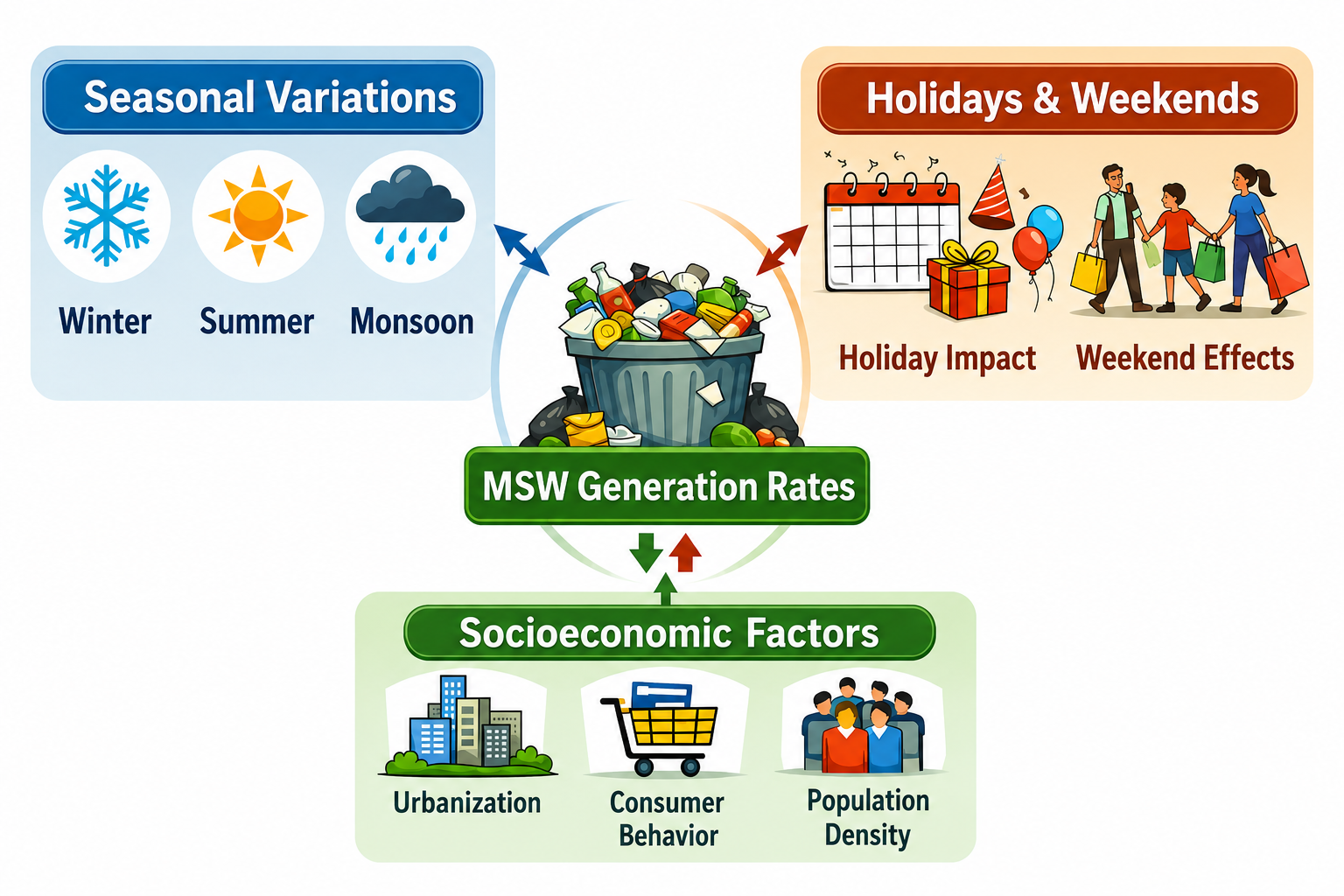}
		\caption{Factors affecting MSW generation }
		\label{MSW}
	\end{figure}

	Trash management has emerged as a critical concern in urban areas across the globe which triggered in a continuous escalation in MSW formation \cite{dey2021smart, deybookchapter}. Recent innovations in IoT-powered smart trash handling and deep learning dependent waste classification systems have improved trash monitoring, segregation, and collection efficiency \cite{wang2021smart}. MSW generation rates are shaped by multiple interacting parameters \cite{alzamora2022forecasting}. Temporal factors such as weekdays, weekends, and holidays introduce short-term fluctuations due to variations in human activities and consumption behavior. Seasonal variations (Winter, Summer, Monsoon, and Post-monsoon) capture recurring patterns driven by climate dependent lifestyle changes. Additionally, socioeconomic and behavioral factors including urbanization, consumption intensity, population density, and economic conditions govern long-term waste generation trends. Figure \ref{MSW} shows different parameters that affect MSW generation.
	
	Effective waste management relies on the accurate prediction of MSW generation rates, as it enables the optimization of collection schedules, allocation of resources such as collection vehicles or waste bins, and the formulation of sustainable waste disposal strategies. Smart Waste Management frameworks \cite{deybookchapter, dey2021smart, dey2022covid} have been proposed to address these challenges by integrating intelligent data-driven solutions. However, accurate forecasting of MSW generation remains difficult due to the interdependent spatial and temporal dependencies and the influence of seasonal and socio-economic factors.
	
	Time series analysis \cite{9184830} such as LSTM \cite{dey2024waste, ahmed2022forecasting, nourani2025predicting, mudannayake2022exploring}, Bidirectional LSTM (BLSTM) \cite{jammeli2021sequential}, nonlinear Autoregressive (AR) models \cite{kumar2021forecasting}, multi-site LSTM \cite{cubillos2020multi}, Recurrent Neural Network (RNN) \cite{vu2022analysis}, Prophet \cite{dey2024waste, mudannayake2022exploring}, SARIMAX \cite{dey2024waste}, Auto ARIMA \cite{mudannayake2022exploring}, Temporal Convolutional Network (TCN) \cite{mudannayake2022exploring}, Transformer\cite{mudannayake2022exploring}, Neural Basis Expansion Analysis for Time Series(N-BEATS) \cite{mudannayake2022exploring} help capture trends, seasonal variations, and temporal dependencies in historical waste data, thereby improving the accuracy of waste generation forecasts.
	
	Despite these advancements, existing works often suffer from limitations such as the unavailability of granular data from lower administrative units \cite{jammeli2021sequential} or small dataset sizes that hinder long-term prediction accuracy \cite{lin2023data,ahmed2022forecasting, jammeli2021sequential, kumar2021forecasting}. Moreover, many of these studies overlook essential factors such as seasonality and holiday effects \cite{vu2022analysis,cubillos2020multi, jammeli2021sequential, nourani2025predicting, mudannayake2022exploring}, which significantly influence daily waste generation patterns.
	
	To overcome these limitations, in this work, we consider a real-world dataset of Sri Lanka \cite{kaggle_dataset} and consider several impacting parameters, such as holidays, weekend-weekdays, seasons, temperature, and humidity, that can influence the MSW generation rate. We implement four widely used time series forecasting models, Prophet, SARIMAX, LSTM and GRU, to check the accuracy of prediction with respect to different parameters. 
	
	The initial version of the work has been published in a conference \cite{dey2024waste}. This is the extension of that work. The major contributions are as follows:

%
%
%
%

	(1) An enhanced dataset \cite{msw_extended_dataset} is constructed by integrating temperature, humidity, and holiday flags, into the original dataset \cite{kaggle_dataset}, to better capture seasonal and temporal dynamics in MSW generation. 
	
	(2) The comparative analysis is expanded by incorporating an additional deep learning model GRU, alongside the previously considered LSTM, SARIMAX, and Prophet models. 
	
	(3) An extensive category-wise evaluation is performed to analyse model behaviour. This includes holidays and non-holidays, weekdays and weekends. Also, the evaluation considers seasonal periods like winter, summer, monsoon, and post monsoon. It also considers different temperature and humidity ranges. This multi-dimensional assessment provides deeper insight into how different factors influence MSW generation.
	
	(4) This work evaluates the efficiency of the models through standard error metrics such as MAPE, MAE, RMSE and $R^2$.

	The organization of the remaining sections of this work is outlined as: Section~\ref{RW} reviews the related works that use time series forecasting models in the domain of MSW forecasting. Section~\ref{gen} presents the four models used. Section ~\ref{ex} discusses the experimental setup. Section \ref{result} presents the comparative performance analysis of the considered models. Finally, Section~\ref{con} concludes the work and presents opportunities for subsequent research.

	\section{Related Work}
\label{RW}

The present section discusses the state-of-the-art of the time-series forecasting methods used for predicting MSW production.

Authors of \cite{ahmed2022forecasting} examines the influence of lifestyle variations on waste composition in East Cairo, Egypt, using LSTM-based forecasting on real datasets comprising only 100 days of observations. Authors of \cite{jammeli2021sequential} applies LSTM and BLSTM to forecast monthly MSW generation and estimate the required number of waste bins using monthly MSW data for only one year (January–December 2017) collected from the city of Sousse, Tunisia. In \cite{cubillos2020multi}, a LSTM network is utilized to forecast household waste generation using data from Herning Municipality, Denmark. However, these three works do not incorporate either seasonality or holiday effects, which may limit the ability to capture realistic temporal variations in MSW generation.

Authors of \cite{nourani2025predicting} utilize Feed Forward Neural Network(FFNN) and LSTM models with Mutual Information(MI) with Shapley Additive Explanations (SHAP) based feature selection for MSW generation prediction by incorporating meteorological and demographic variables such as temperature, precipitation, income, population and Consumer Price Index (CPI).
Authors of \cite{kumar2021forecasting} predict monthly MSW generation in Nagpur, India using Non-linear Autoregressive (NAR) models combined with classical decomposition and LR, where the data span is limited to April 2015–March 2017. Authors of \cite{vu2022analysis} demonstrate the capability of k-fold cross-validation by developing seven RNN–LSTM models for weekly landfill waste disposal prediction in Regina, Canada, using weekly data from 2013–2019. The work incorporates climatic variables including temperature, humidity and wind speed, along with economic determinants like income and unemployment levels. Model performance is evaluated through a 7-fold cross-validation framework with a one-year testing set in each fold. In these three works holiday effects are not explicitly considered in the forecasting framework, potentially limiting realistic modeling of temporal fluctuations in MSW generation.

Authors of \cite{lin2023data} apply Prophet combined with Household Hazardous Waste Visual Geometry Group Networks (HVGGNet) for Household Hazardous Waste forecasting and classification in Shanghai using observations collected from March 2020 to December 2020. This work explicitly modeles trend, seasonality and holiday effects through the Prophet framework. However, the relatively short observation period may limit robust learning of long-term seasonal and holiday-related patterns.

Authors of \cite{mudannayake2022exploring} explore multiple ML and DL models, including Prophet, LSTM, TCN, Transformer and N-BEATS, for multi-step MSW generation forecasting using datasets from Sri Lanka, Austin, and Ballarat. Although seasonality is incorporated for the three datasets and holiday effects are considered for the Austin and Ballarat datasets only, but not incorporated for the Sri Lankan dataset.

A common limitation across these works is that none of them comprehensively investigate the influence of weekday–weekend patterns on municipal solid waste generation behaviour. Table~\ref{tab:relatedwork} discusses the comparative analysis of the existing works.

\begin{table}[t]
	\centering
	\caption{Comparative Assessment of existing works on MSW generation prediction using Time Series Forecasting}
	\label{tab:relatedwork}
	
	\scriptsize
	\renewcommand{\arraystretch}{0.9}
	
	\resizebox{\textwidth}{!}{%
		
		\begin{tabular}{
				>{\raggedright\arraybackslash}p{1.6cm}
				>{\raggedright\arraybackslash}p{2.8cm}
				>{\raggedright\arraybackslash}p{2.4cm}
				>{\raggedright\arraybackslash}p{2.3cm}
				>{\centering\arraybackslash}p{1.1cm}
				>{\centering\arraybackslash}p{1.1cm}
				>{\centering\arraybackslash}p{1.3cm}
				>{\centering\arraybackslash}p{1.2cm}
				>{\centering\arraybackslash}p{1.2cm}
				>{\raggedright\arraybackslash}p{2.6cm}
			}
			
			\hline

			\textbf{Ref} &
			\textbf{Method Used} &
			\textbf{Area} &
			\textbf{Data} &
			\textbf{Seasonality} &
			\textbf{Holiday} &
			\textbf{Weekday vs Weekend} &
			\textbf{Temperature effect} &
			\textbf{Humidity effect} &
			\textbf{Data Limitations}
			\\
			
			\hline

			\cite{ahmed2022forecasting}
			& LSTM
			& East Cairo
			& 100 days
			& \cross
			& \cross
			& \cross
			& \cross
			& \cross
			& Very short duration data
			\\
			
			\cite{jammeli2021sequential}
			& LSTM, BLSTM
			& Tunisia
			& 2017 y
			& \cross
			& \cross
			& \cross
			& \cross
			& \cross
			& Single year data
			\\
			
			\cite{kumar2021forecasting}
			& NAR Neural Network
			& India (Nagpur)
			& 2015--2017 y
			& \tick
			& \cross
			& \cross
			& \cross
			& \cross
			& Limited data
			\\
			
			\cite{cubillos2020multi}
			& Multi-site LSTM
			& Denmark
			& 2011--2018 y
			& \cross
			& \cross
			& \cross
			& \cross
			& \cross
			& -
			\\
			
			\cite{vu2022analysis}
			& RNN--LSTM
			& Canada
			& 2013--2019 y
			& \tick
			& \cross
			& \cross
			& \tick
			& \tick
			& -
			\\
			
			\cite{lin2023data}
			& Prophet + HVGGNet
			& Shanghai
			& 2020 y
			& \tick
			& \tick
			& \cross
			& \cross
			& \cross
			& Limited data
			\\
			
			\cite{nourani2025predicting}
			& FFNN + LSTM + MI--SHAP
			& Austin (USA), Ballarat (Australia), Boralesgamuwa (Sri Lanka)
			& 2005--2020 y, 2001--2014 y, 2012--2018 y respectively
			& \tick
			& \cross
			& \cross
			& \tick
			& \cross
			& -
			\\
			
			\cite{mudannayake2022exploring}
			& Linear Regression, Auto ARIMA, LightGBM, Random Forest, Prophet, LSTM, TCN, Transformer, N-BEATS
			& Ballarat (Australia), Austin (USA), Dehiwala (Sri Lanka), Boralesgamuwa (Sri Lanka), Moratuwa (Sri Lanka)
			& 2000--2015 y, 2003--2021 y, 2012--2018 y, 2012--2018 y, 2014--2018 y
			& \tick
			& \tick
			& \cross
			& \cross
			& \cross
			& -
			\\
			
			\hline
			
		\end{tabular}
		
	}
	
\end{table}

In our proposed work, we systematically address the key gaps identified in existing MSW generation prediction studies. To overcome the limitations associated with small-scale datasets, a real-world, multi-year (2012–2018) MSW dataset from Sri Lanka is used. We apply and compare four widely used time series models, namely, Prophet, SARIMAX, LSTM and GRU. 
We explicitly analyze the holidays, weekday–weekend, seasonal divisions (Winter, Summer, Monsoon, and Post-monsoon), temperature and humidity effects into the forecasting models to capture real-world waste generation dynamics. 
This comprehensive framework contributes to improving the reliability and practical applicability of MSW prediction.

	\section{Background of Forecasting Models}
\label{gen}

Four models are applied in this paper for MSW generation prediction, namely Prophet, SARIMAX, LSTM and GRU. These models are selected because they have been extensively used in time-series forecasting applications \cite{lin2023data, banas2021evaluating, mudannayake2022exploring, xiong2024time}. 
The following subsections briefly describe the theoretical foundations of each model.

\subsection{Prophet}

Prophet is designed by Facebook which is an open source model constructed for time series forecasting using an additive approach \cite{taylor2018forecasting}. It demonstrates good performance in dealing with outliers and missing observations \cite{rostami2021forecasting}. The model shows time series as a combination of four components, expressed as follows: $y(t) = g(t) + s(t) + h(t) + \epsilon(t)$ where $y(t)$ is the value being predicted at each timestamp, $g(t)$ represents trend, $s(t)$ depicts seasonality, $h(t)$ defines holiday and $\epsilon(t)$ is the error term.

\subsection{LSTM}

The Long Short Term Memory (LSTM) network \cite{weerakody2023cyclic} is an enhanced form of a Recurrent Neural Network (RNN) designed to retain information over time. Each LSTM cell contains a memory unit that preserves previously acquired information, updates its contents when required and transfers relevant information to subsequent cells or layers. The memory mechanism is regulated by three gates: the input gate, the output gate, and the forget gate. 


\textbf{Input Gate:} The input gate regulates the information to be incorporated into the cell state. A sigmoid activation function determines which input data should be updated and stored in the memory unit. The information is then integrated into the cell state with the help of an activation function. The input gate is mathematically expressed as follows: $I_t = \sigma(W_I [H_{t-1}, x_t] + b_I)$, where, at time $t$, $I_t$ denotes the input gate activation, $\sigma$ represents the sigmoid activation function, $W_I$ is the input gate weight matrix, $H_{t-1}$ represents the hidden state at the previous time step, $x_t$ denotes the input vector for the present time step, and $b_I$ is the input gate bias term.

\textbf{Output Gate:} The output gate controls the information transmitted from the LSTM cell to the subsequent layer for the present time step. The sigmoid activation function determines which components of the cell state $C_t$ should help produce the output. The output gate is mathematically expressed as follows: $O_t = \sigma\left(W_O [H_{t-1}, x_t] + b_O\right)$; where, at time $t$, $O_t$ denotes the output gate activation, $W_O$ represents the output gate weight matrix, and $b_O$ denotes the output gate bias term. The remaining parameters are defined earlier.

\textbf{Forget Gate:} The forget gate regulates the retention and removal of information from the previous cell state. It determines which information should be preserved and which can be discarded before passing the updated information to subsequent processing steps. The forget gate is mathematically represented as follows: $F_t = \sigma(W_F [H_{t-1}, x_t] + b_F)$, where, at time $t$, $F_t$ denotes the forget gate activation vector, $W_F$ represents the forget gate weight matrix, and $b_F$ denotes the forget gate bias term.

\subsection{GRU}
The Gated Recurrent Unit (GRU) is a refined form of the LSTM method, designed to manage information flow using two main gates: the update gate(\( z_t \)) and the reset gate (\( r_t \)) \cite{weerakody2023cyclic}. The process is initialized from a starting point \( z_0 \) and for each time step \( t \), takes the input vector $x_t$ along with the preceding hidden state $h_{t-1}$ to compute the new hidden state $h_t$. The computations, which involve both the reset and update gates feeding into a tanh activation function to produce the candidate state, are defined as follows:
\[
\begin{aligned}
	r_t &= \sigma(W_{xr} x_t + W_{hr} h_{t-1} + b_r), \\
	z_t &= \sigma(W_{xz} x_t + W_{hz} h_{t-1} + b_z), \\
	\tilde{h}_t &= \tanh(W_{xh} x_t + W_{hh} (r_t \odot h_{t-1}) + b_h), \\
	h_t &= z_t \odot h_{t-1} + (1 - z_t) \odot \tilde{h}_t.
\end{aligned}
\]
In these equations, the reset gate $r_t$ determines how much of the preceding hidden state $h_{t-1}$ is used in computing the candidate activation $h_t$. The update gate $z_t$ decides the balance between preserving the previous state $h_{t-1}$ and incorporating the new candidate state $h_t$. By consolidating the input and forget gates of the LSTM into a single update mechanism and omitting an explicit memory cell, the GRU achieves performance on par with LSTMs while offering greater computational efficiency. This architecture enables GRUs to effectively model short term along with long term dependencies within sequential data \cite{sciencedirectGRU}, \cite{mulahuwaish2022covidmis20}.


%
%

\subsection{SARIMAX}
SARIMAX combines Seasonal Autoregression (SAR), Seasonal Integration (SI), and Seasonal Moving Average (SMA) components, depicted by $P$, $D$, and $Q$, respectively. It extends the SARIMA model by incorporating one or more exogenous variables to investigate the effects of external factors on the forecasted time series \cite {kim2022improved}. The SARIMAX model is represented as:

$\phi_p(L)\Phi_P(L^s)(1-L)^d(1-L^s)^D Y_t
=
\theta_q(L)\Theta_Q(L^s)\varepsilon_t
+
\sum_{j=0}^{r}\beta_j X_{t-j}$

where $Y_t$ represents the time series value at time $t$, $L$ is the lag operator such that $L^k Y_t = Y_{t-k}$, and $s$ denotes the seasonal period. $p$, $d$, and $q$ denote the orders of the non-seasonal autoregressive, differencing, and moving average components respectively. $P$, $D$, and $Q$ are the orders of the seasonal autoregressive, differencing and moving average components respectively, while $s$ denotes the seasonal period. The term $X_{t-j}$ represents the exogenous variable at lag $j$, and $\beta_j$ is its corresponding coefficient. The term $\varepsilon_t$ represents the error term, while $\phi_p(L)$, $\theta_q(L)$, $\Phi_P(L^s)$, and $\Theta_Q(L^s)$ are the non-seasonal and seasonal autoregressive and moving average polynomials respectively \cite{banas2021evaluating}.

\section{Experimental Setup}
\label{ex}
This section presents the experimental configuration used for MSW generation forecasting using the four predefined models. It describes the overall experimental framework, preparation of dataset, data preprocessing steps, parameter settings of the different forecasting models and the evaluation criteria used in this work. 

\subsection{Environment}

All experiments are implemented in Python and executed using the Google Colab environment, which provides GPU acceleration for computationally intensive deep learning models. The development, data preprocessing, and preliminary analyses are carried out on a MacBook Pro integrated with an Apple M2 processor, 8 GB unified memory, and running macOS Sequoia (Version 15.6). The forecasting models, i.e. Prophet, SARIMAX, LSTM, and GRU, are implemented using the Prophet, Statsmodels, TensorFlow, and Keras libraries respectively. Data preprocessing and visualization are performed using Pandas, NumPy, and Matplotlib.
	\subsection{Dataset Preparation}

The dataset for Boralasgamuwa is obtained from \cite{kaggle_dataset}, covering the period from 2012 to 2018. The original dataset includes the attributes \textit{area}, \textit{ticket\_date}, \textit{waste\_type}, and \textit{net\_weight\_kg}.

To enhance the accuracy of waste generation prediction, additional environmental attributes such as \textit{temperature} and \textit{humidity} are incorporated, as these factors are known to influence waste generation patterns \cite{zhang2022machine}. 
The temperature and humidity are calculated from the Weather Underground historical records available on Weather Underground \cite{wunderground}.

In addition, Sri Lankan national holidays \cite{holiday} are encoded as a binary flag (1 for holiday, 0 otherwise) to capture variations in waste generation. 
After integrating these additional features, the final dataset contains \textit{area}, \textit{ticket\_date}, \textit{waste\_type}, \textit{net\_weight\_kg}, \textit{temperature (C)}, \textit{humidity (\%)}, and holiday as shown at \cite{msw_extended_dataset}.


%
%

%
%

\subsection{Data Preprocessing}
The steps of data preprocessing are illustrated below:

\textbf{Date Completion and Missing Value Handling:}

Dates absent from the original dataset are introduced as NaN values during reindexing, which are subsequently handled using the built-in linear interpolation functionality provided by the Pandas library. This interpolation process estimates the missing waste values based on neighbouring observations, thereby maintaining a continuous time series for subsequent modelling.

\textit{Example:} In the dataset \cite{msw_extended_dataset}, the value for 05/11/2015 was missing. The built-in linear interpolation estimated the missing value as 44,075 kg based on these neighbouring observations on 04/11/2015 (38,520 kg) and 06/11/2015 (49,630 kg).

\medskip

\textbf{Anomaly Detection:}

Anomalies are the values which have substantial deviation from neighbouring observations. Anomalies are detected using a double-stage Isolation Forest approach to enhance the robustness of outlier identification.  

\textit{Example:} For instance, the waste generation values on 11/08/2015, 12/08/2015, 14/08/2015, and 15/08/2015 were 17,240 kg, 37,850 kg, 46,220 kg, and 32,650 kg, respectively, which are substantially lower than the observed value on 13/08/2015 (159,020 kg). That's why it is consudered as anomaly.  

\medskip

 \textbf{Anomaly Imputation:}  
Detected anomalous observations are imputed using a strategy comprising lag-based substitution to preserve weekly seasonality.
	
	\textit{Example:} The observation on 13/08/2015 (159,020 kg) is identified as anomalous. It is replaced using the corresponding 7-day lagged observation from 06/08/2015 (29,670 kg).

\subsection{Model Parameters}
In the proposed approach, the daily MSW prediction performance is evaluated using Prophet, SARIMAX, LSTM and GRU time series models. These models applied on a real-world Sri Lankan dataset covering a substantial time span (2012 to 2018) \cite{msw_extended_dataset}.  Training period is the year 2012-2017 and testing period is 2018.

Table \ref{tab:model_parameters} summarizes the parameter settings used to implement the Prophet, LSTM, SARIMAX, and GRU forecasting models.

\begin{table}
	\centering
	\caption{Parameter settings of the four forecasting models.}
	\label{tab:model_parameters}
	\scriptsize
	
	\begin{minipage}{0.48\linewidth}
		\centering
		\textbf{(a) Prophet Model}
		\vspace{2pt}
		
		\begin{tabular}{|p{4cm}|p{3.1cm}|}
			\hline
			\textbf{Parameter} & \textbf{Value / Setting} \\
			\hline
			\texttt{changepoint\_prior\_scale} & 0.8 \\
			\hline
			\texttt{seasonality\_prior\_scale} & 25 \\
			\hline
			\texttt{holidays\_prior\_scale} & 15 \\
			\hline
			\texttt{seasonality\_mode} & additive \\
			\hline
			\texttt{custom\_seasonality\_1} & monthly (period = 30.5, fourier\_order = 20) \\
			\hline
			\texttt{custom\_seasonality\_2} & weekly (period = 7, fourier\_order = 10) \\
			\hline
		\end{tabular}
	\end{minipage}
	\hfill
	\begin{minipage}{0.48\linewidth}
		\centering
		\textbf{(b) LSTM Model}
		\vspace{2pt}
		
		\begin{tabular}{|p{2.4cm}|p{3.2cm}|}
			\hline
			\textbf{Parameter} & \textbf{Value / Setting} \\
			\hline
			Sequence Length & 60 \\
			\hline
			LSTM Layers & 128-64-32 units \\
			\hline
			Dropout & 0.2 after first two LSTM layers \\
			\hline
			Optimizer & Adam \\
			\hline
			Learning Rate & 0.0005 \\
			\hline
			Maximum Epochs & 100 \\
			\hline
			Early Stopping Patience & 15 epochs \\
			\hline
			Batch Size & 16 \\
			\hline
			Random Seed & 42 \\
			\hline
		\end{tabular}
	\end{minipage}
	
	\vspace{0.5cm}
	
	\begin{minipage}{0.48\linewidth}
		\centering
		\textbf{(c) SARIMAX Model}
		\vspace{2pt}
		
		\begin{tabular}{|p{3.5cm}|p{1.6cm}|p{2.0cm}|}
			\hline
			\textbf{Parameter} & \textbf{Value} & \textbf{Description} \\
			\hline
			\texttt{order} & (1,1,1) & $p=1$ AR, $d=1$ differencing, $q=1$ MA \\
			\hline
			\texttt{seasonal\_order} & (1,1,1,7) & $P=1$ seasonal AR, $D=1$ seasonal differencing, $Q=1$ seasonal MA, $s=7$ days \\
			\hline
			\texttt{trend\_decomposition} & Additive, period=7 & Extracts trend and weekly seasonality \\
			\hline
			\texttt{stationarity\_check} & ADF ($p=0.0080$) & Confirms stationarity ($p<0.05$) \\
			\hline
		\end{tabular}
	\end{minipage}
	\hfill
	\begin{minipage}{0.48\linewidth}
		\centering
		\textbf{(d) GRU Model}
		\vspace{2pt}
		
		\begin{tabular}{|p{2.4cm}|p{3.2cm}|}
			\hline
			\textbf{Parameter} & \textbf{Value / Setting} \\
			\hline
			GRU Layers & 2 (128 units, 64 units) \\
			\hline
			Activation & tanh \\
			\hline
			Dropout & 0.1 \\
			\hline
			Optimizer & Adam (learning rate = 0.001) \\
			\hline
			Batch Size & 16 \\
			\hline
			Epochs & 200 (with early stopping) \\
			\hline
			Early Stopping Patience & 15 \\
			\hline
			Sequence Length & 30 days \\
			\hline
		\end{tabular}
	\end{minipage}
	
\end{table}

\subsection{Evaluation Metrics}
Forecast performances of all models are evaluated using RMSE, MAE, MAPE and $R^2$ which are described next.

\subsubsection{Root Mean Squared Error (RMSE)} 	
Root Mean Squared Error (RMSE) is a statistical metric used to quantify the magnitude of prediction errors by measuring the differences between actual and predicted values \cite{jierula2021study}. It is calculated using the following expression: $$ \text{RMSE} = \sqrt{\frac{1}{n}\sum_{i=1}^{n}(y_i - \hat{y}_i)^2}$$
Here, $n$ denotes the total number of observations, $y_i$ represents the actual value of the $i$-th observation, and $\hat{y}_i$ denotes its equivalent predicted value. A lower RMSE reflects that the predicted values are closer to the actual observations which reflecting better forecasting accuracy.

\subsubsection{Mean Absolute Error (MAE)}

Mean Absolute Error (MAE) calculates the average absolute difference between the actual observations and their associated predicted values \cite{jierula2021study}. It is computed using the following equation:
$$\text{MAE} = \frac{1}{n}\sum_{i=1}^{n}|y_i - \hat{y}_i|$$
Here, $n$ specifies the total observation count, $y_i$ denotes the actual value of the $i$-th observation, and $\hat{y}_i$ represents its associated forecasted value. A lower MAE indicates smaller average prediction errors and therefore, better forecasting performance.

\subsubsection{Mean Absolute Percentage Error (MAPE)} 

Mean Absolute Percentage Error (MAPE) evaluates forecasting performance by determining the average absolute percentage deviation between the actual observations and the predicted values for the same observations \cite{jierula2021study}. It is mathematically expressed as follows:
$$	\text{MAPE} = \frac{1}{n}\sum_{i=1}^{n}\left|\frac{y_i - \hat{y}_i}{y_i}\right| \times 100\%$$

Here, $n$ denotes the total number of observations, $y_i$ represents the actual value of the $i$-th observation, and $\hat{y}_i$ indicates its corresponding predicted value. A lower MAPE signifies a smaller percentage deviation between the observed and forecasted values, indicating higher prediction accuracy.

\subsubsection{\texorpdfstring{$R^2$}{R2}}

The coefficient of determination ($R^2$) \cite{kim2020assessing} is a statistical measure used to evaluate how well a prediction model explains the variability in the observed data. It is computed using the following equation: $R^2 = 1 - \frac{\sum_{i=1}^{n} (y_i - \hat{y}_i)^2}{\sum_{i=1}^{n} (y_i - \bar{y})^2}$
Here, $y_i$ denotes the actual value of the $i$-th observation, $\hat{y}_i$ represents the corresponding predicted value, and $\bar{y}$ indicates the mean of all observed values. An $R^2$ value closer to 1 generally indicates a better model fit, whereas lower or negative values suggest poorer predictive performance relative to using the mean of the observed values as the prediction.

\section{Results}
\label{result}

This section analyzes the experimental results by examining the effectiveness of all four forecasting models (Prophet, Sarimax, LSTM, GRU) under multiple influencing factors, including holidays, weekday-weekends, seasonal variations, temperature, and humidity. The analysis highlights how well each model captures temporal patterns, environmental effects, and behavioural variations in waste generation. By examining these diverse conditions, this work provides a detailed understanding of model effectiveness for accurate waste forecasting under varying real-world conditions.

\subsection{Holiday Effect}

This section analyzes the influence of holidays on daily MSW generation and evaluates how the four different forecasting models capture these variations. Holiday periods often exhibit distinct waste generation patterns due to changes in human activity, such as reduced institutional operations or increased social gatherings. To analyze this effect, both per-holiday deviations and comparisons between holiday and non-holiday periods are examined. 

As discussed in section \ref{gen}, Prophet inherently provides a dedicated holiday component and can directly incorporate predefined holiday dates into the forecasting process. In contrast, SARIMAX, LSTM, and GRU do not have an intrinsic holiday component. In this work, for SARIMAX, the holiday flag is incorporated as an exogenous variable, while for LSTM and GRU holiday is included as an additional input feature during model training. 

\begin{figure}
	\centering
	
	\begin{subfigure}{0.48\linewidth}
		\centering
		\includegraphics[width=\linewidth]{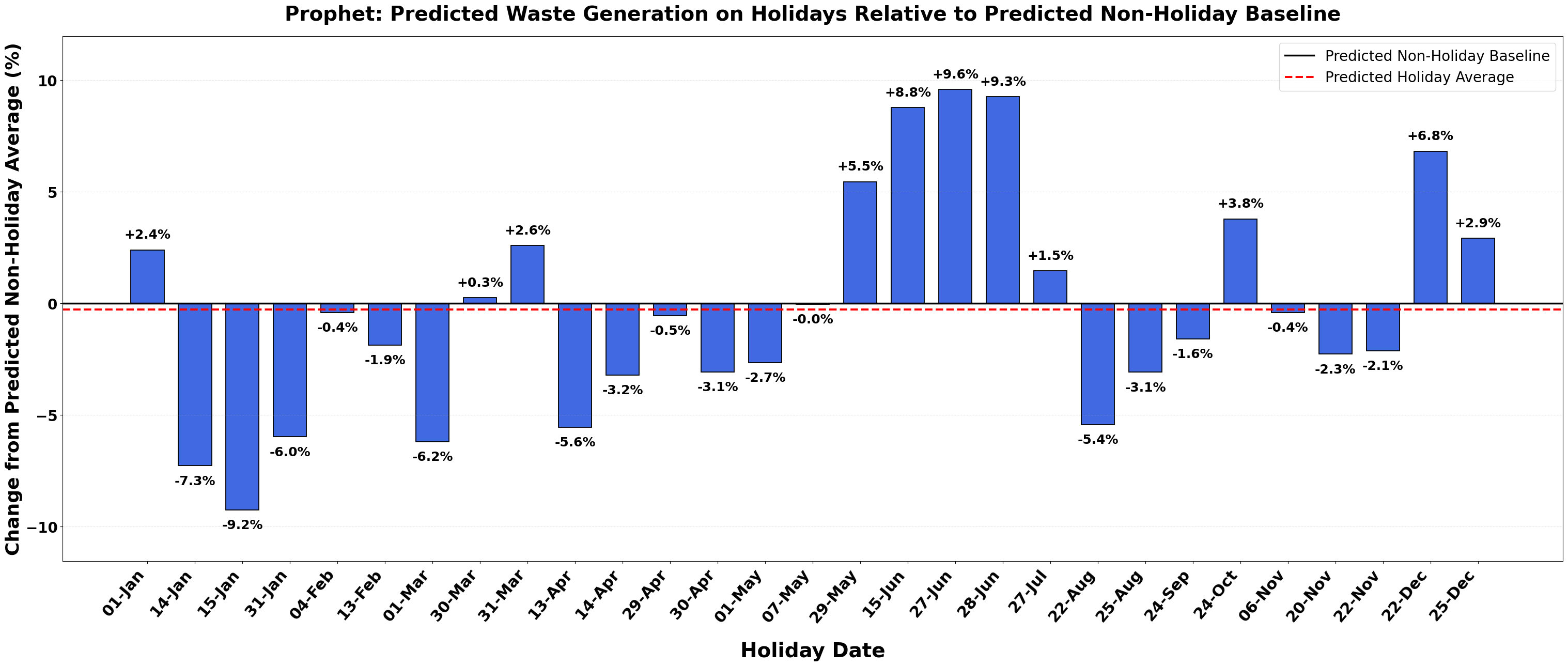}
		\caption{Prophet}
		\label{fig:holiday_prophet}
	\end{subfigure}
	\hfill
	\begin{subfigure}{0.48\linewidth}
		\centering
		\includegraphics[width=\linewidth]{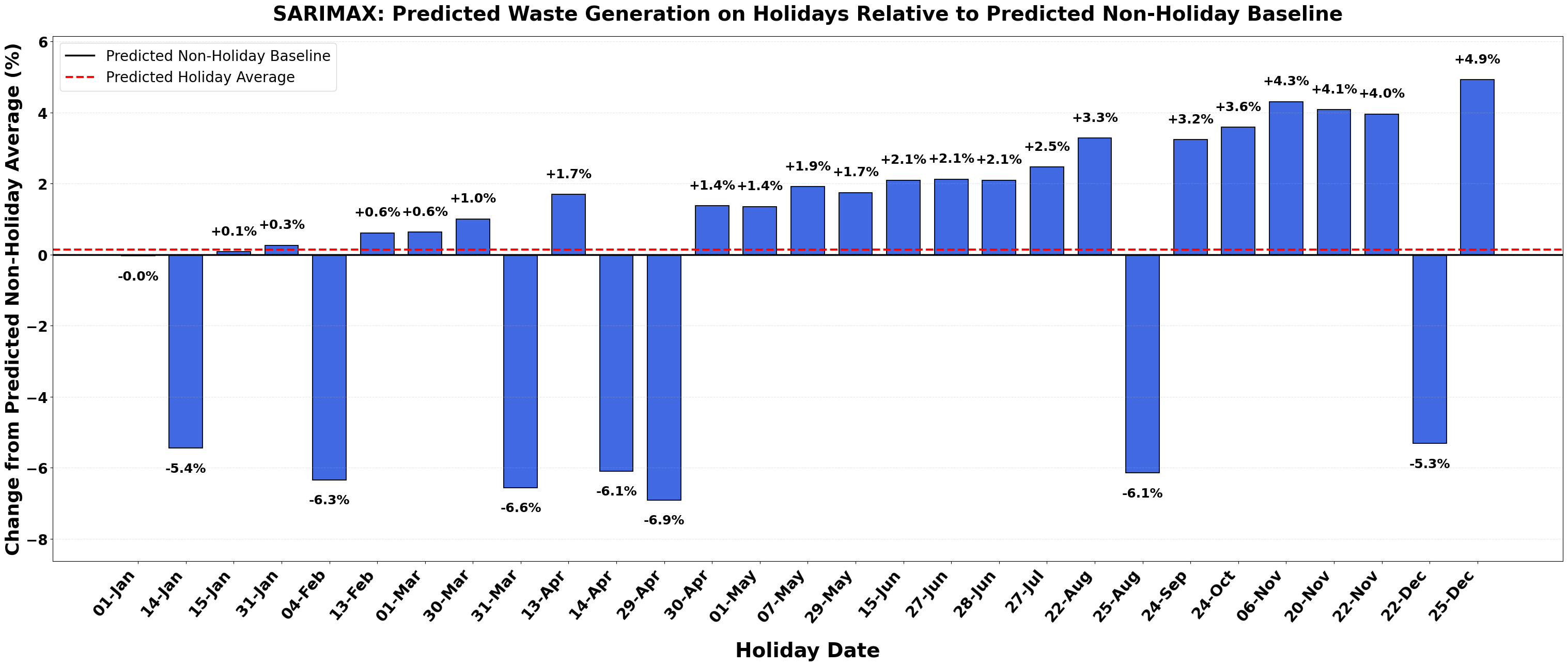}
		\caption{SARIMAX}
		\label{fig:holiday_sarimax}
	\end{subfigure}
	
	\vspace{0.3cm}
	
	\begin{subfigure}{0.48\linewidth}
		\centering
		\includegraphics[width=\linewidth]{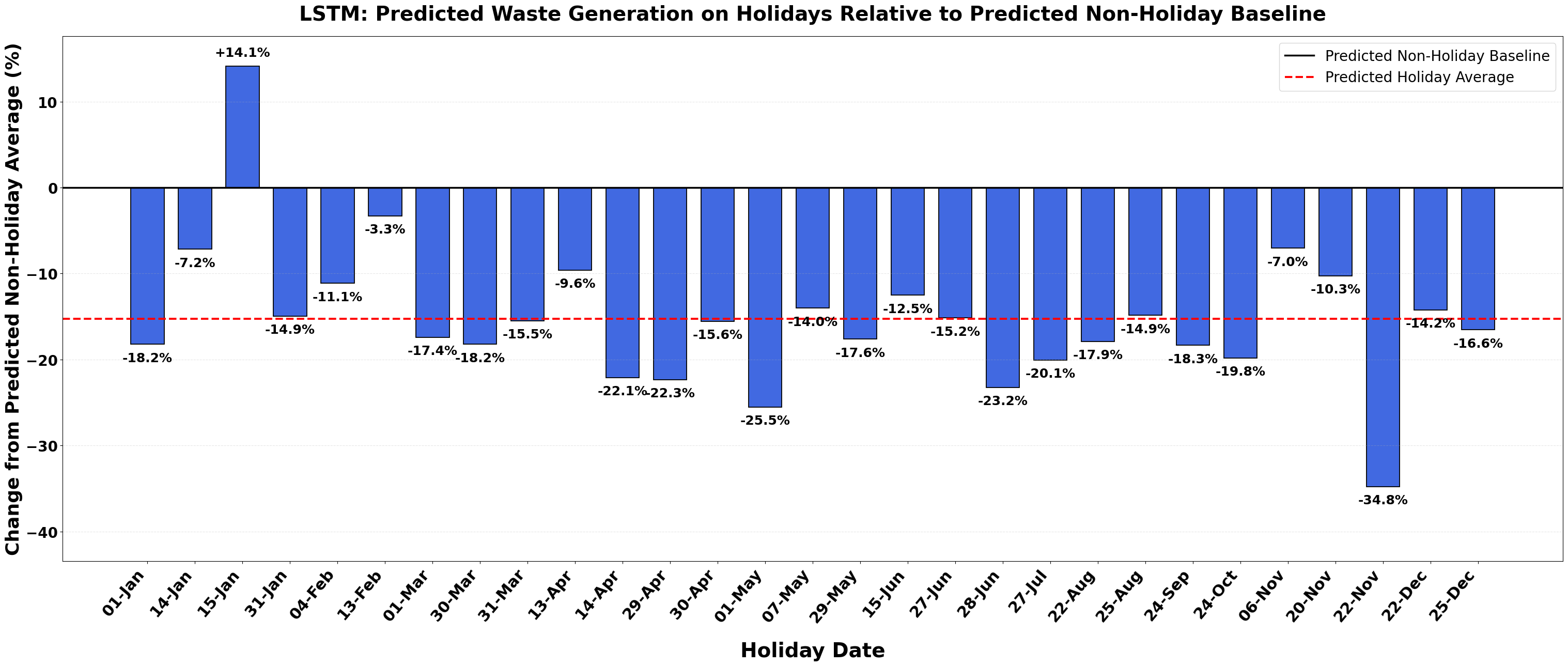}
		\caption{LSTM}
		\label{fig:holiday_lstm}
	\end{subfigure}
	\hfill
	\begin{subfigure}{0.48\linewidth}
		\centering
		\includegraphics[width=\linewidth]{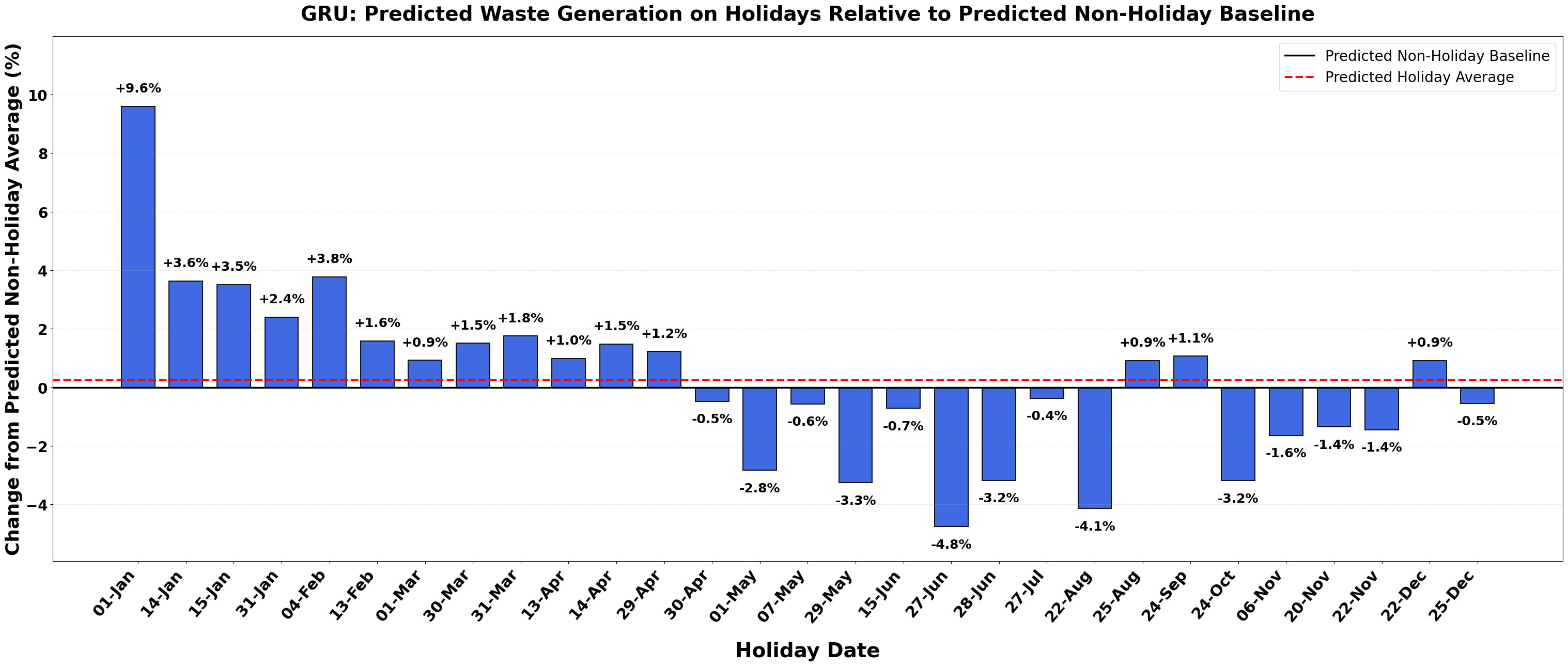}
		\caption{GRU}
		\label{fig:holiday_gru}
	\end{subfigure}
	
	\vspace{0.3cm}
	
	\begin{subfigure}{0.90\linewidth}
		\centering
		\includegraphics[width=\linewidth]{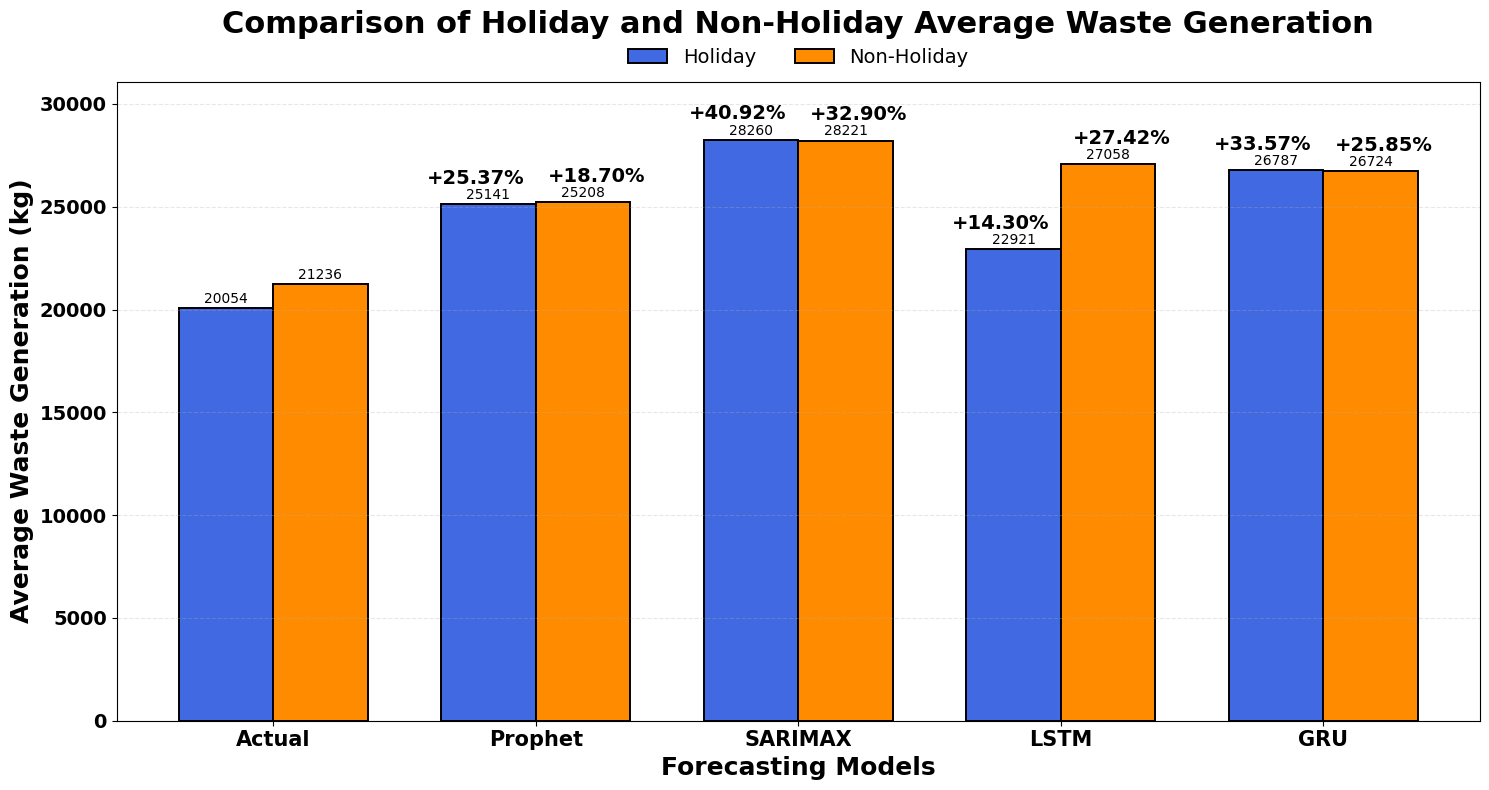}
		\caption{Actual and predicted average waste generation during holiday and non-holiday periods}
		\label{fig:holiday_nonholiday_merged}
	\end{subfigure}
	
	\caption{Per-holiday MSW production change predicted by all four models and comparison of actual and predicted average waste generation during holiday and non-holiday periods.}
	\label{fig:holiday_comparison}
\end{figure}

Figures~\ref{fig:holiday_prophet}, \ref{fig:holiday_sarimax}, \ref{fig:holiday_lstm}, and \ref{fig:holiday_gru}  illustrate the model-wise (Prophet, SARIMAX, LSTM, and GRU, respectively) predicted waste generation on individual holidays relative to the corresponding model-predicted non-holiday baseline in 2018. Here, the black horizontal reference line represents the predicted average waste generation during non-holiday periods for each model and serves as the baseline for calculating the percentage change. The individual bars indicate how much higher or lower (in \%) the predicted waste generation is for each holiday compared with the baseline. The red dotted line represents the corresponding model's average predicted waste generation during holidays. 
The results show considerable differences in the predicted holiday effects across the four models. Prophet and LSTM predicts a decrease of 0.26\% an 15.29\% in waste generation during holidays relative to its predicted non-holiday baseline. SARIMAX predicts an increase of 0.14\%, while GRU predicts an increase of 0.23\%.

%
%
%
%

Fig.~\ref{fig:holiday_nonholiday_merged} presents a comparison of the actual and model-predicted average waste generation during holiday and non-holiday periods in 2018. The actual observations show that waste generation decreases during holidays, with the average decreasing from 21,236 kg during non-holidays to 20,054 kg during holidays, corresponding to a reduction of approximately 5.56\%. Among the forecasting models, Prophet and LSTM reproduce this observed direction, predicting lower waste generation during holidays than during non-holidays. In contrast, SARIMAX and GRU predict slightly higher waste generation during holidays than during non-holidays. Considering the average prediction deviation, LSTM achieves the lowest deviation (20.86\%), compared with 22.04\% for Prophet, and therefore provides the best overall prediction performance among the evaluated models. 

%
%
%
%


\subsection{Weekday and Weekend Effect}

This section presents a comparative analysis of municipal solid waste generation patterns between weekdays and weekends using all the four forecasting models, aiming to capture and understand model-specific representations of weekly waste dynamics.

To explicitly account for the different waste-generation patterns associated with weekdays and weekends, the historical dataset from 2012-2017 is divided into two independent training subsets based on the day of the week. Observations from Monday to Friday are used exclusively to train the weekday model, while observations from Saturday and Sunday are used exclusively to train the weekend model. 

\begin{figure}
	\centering
	
	\begin{subfigure}{0.48\linewidth}
		\centering
		\includegraphics[width=\linewidth]{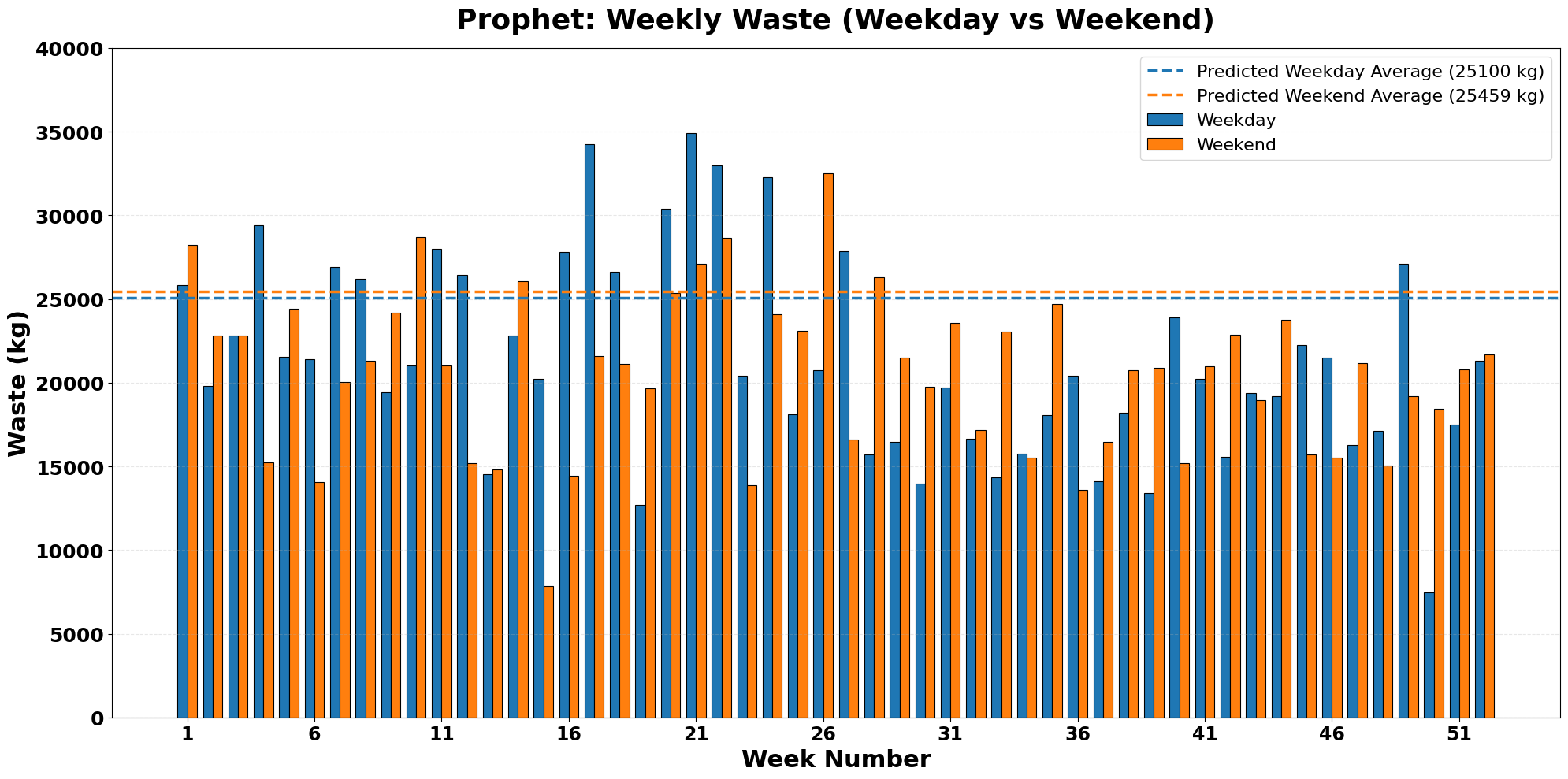}
		\caption{Prophet}
		\label{fig:week_prophet}
	\end{subfigure}
	\hfill
	\begin{subfigure}{0.48\linewidth}
		\centering
		\includegraphics[width=\linewidth]{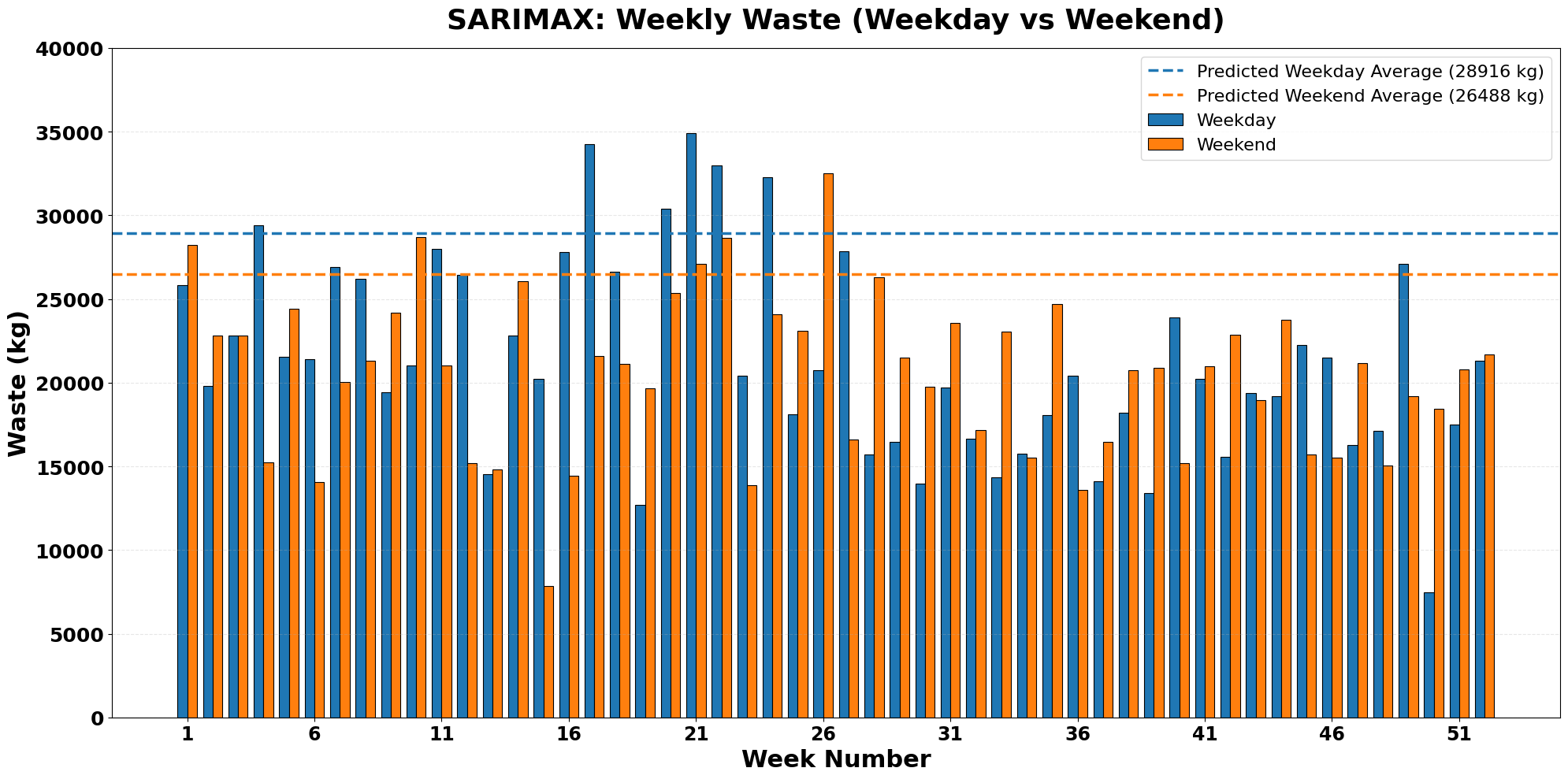}
		\caption{SARIMAX}
		\label{fig:week_sarimax}
	\end{subfigure}
	
	\vspace{0.3cm}
	
	\begin{subfigure}{0.48\linewidth}
		\centering
		\includegraphics[width=\linewidth]{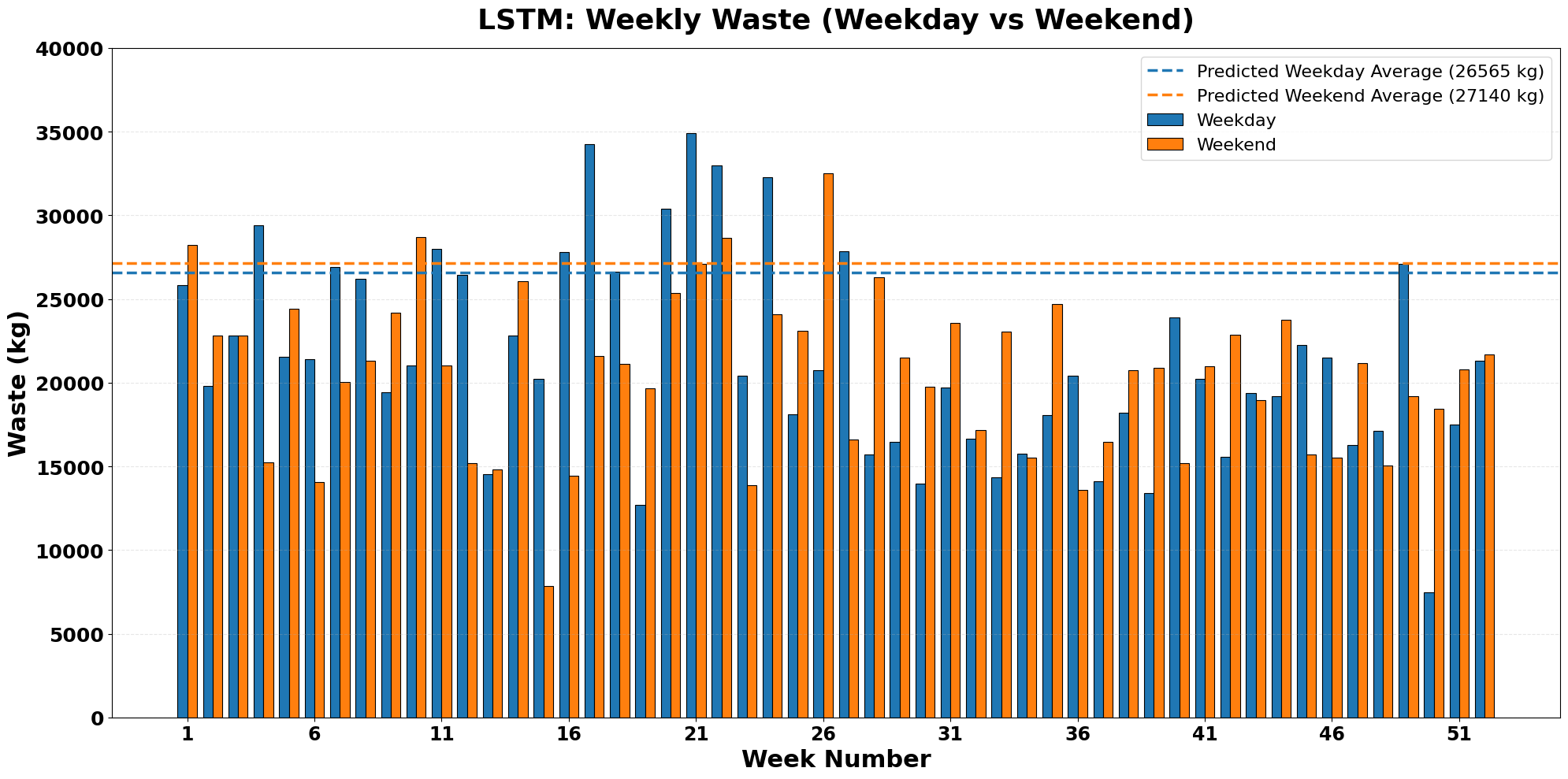}
		\caption{LSTM}
		\label{fig:week_lstm}
	\end{subfigure}
	\hfill
	\begin{subfigure}{0.48\linewidth}
		\centering
		\includegraphics[width=\linewidth]{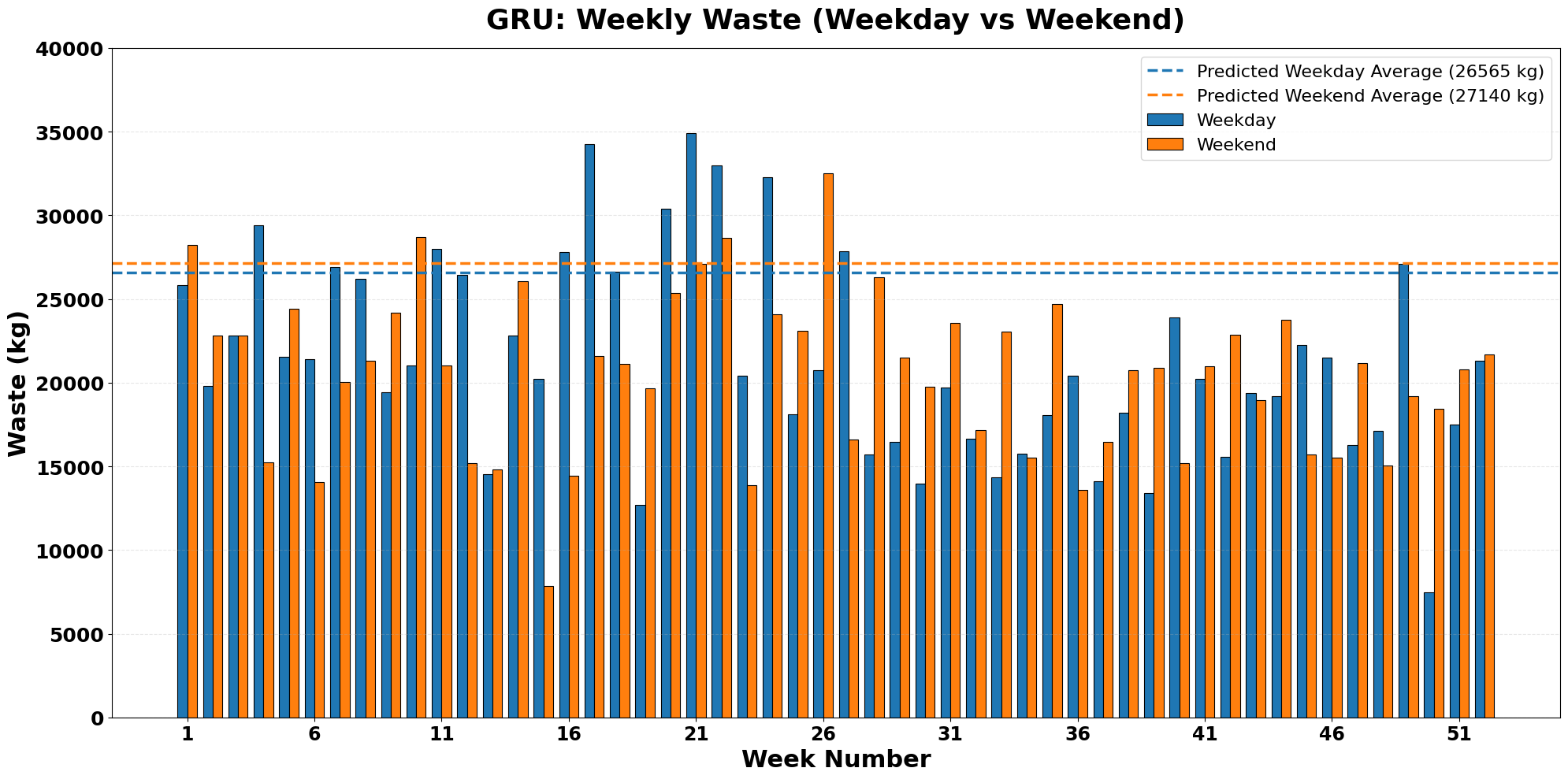}
		\caption{GRU}
		\label{fig:week_gru}
	\end{subfigure}
	
	\caption{Weekday and weekend-wise MSW generation patterns predicted by four forecasting models.}
	\label{fig:weekday_weekend_comparison}
\end{figure}

Figures~\ref{fig:week_prophet}, \ref{fig:week_sarimax}, \ref{fig:week_lstm}, and \ref{fig:week_gru} illustrate tthe weekly predicted waste generation for weekdays and weekends in 2018. The red and purple dotted horizontal lines represent the predicted average waste generation for weekdays and weekends, respectively, over the 2018 forecasting period. SARIMAX predicts a higher average waste generation for weekdays than for weekends. In contrast, Prophet, LSTM and GRU show the opposite pattern, with predicted weekend averages exceeding their weekday averages.

The actual data show a higher average waste generation on weekdays (21385.85 kg) than on weekends (20529.04 kg). Among the forecasting models, SARIMAX is the only model that preserves this weekday-dominant pattern. The stronger weekday-weekend differentiation produced by SARIMAX can be attributed to its explicit weekly seasonal component, which allows recurring seven-day temporal dependencies to be modelled directly. 

%
%
%

\subsection{Seasonal Effect }

This section analyzes the seasonal variation in waste generation and evaluates the performance of different forecasting models across distinguishable seasons. They are Winter (December to February), Summer (March to May), Monsoon (June to September), and Post-Monsoon (October to November). The objective is to examine how effectively each model captures seasonal patterns and fluctuations in waste generation.

Figure~\ref{fig:seasonal_comparison} presents the season-wise comparison of average waste generation prediction across all models. Seasonal variations in waste generation are evident, with the highest average waste generation observed during the Summer season, followed by Winter, while Monsoon and Post-Monsoon exhibit comparatively lower average waste generation. In this figure, positive percentages indicate waste increase, whereas negative percentages indicate waste decrease with respect to the average waste generation in 2018. The 'actual' bar represents the average waste generation for each season in 2018, while the model bar represents the corresponding average forecasted waste generation. 
 Table~\ref{fig:seasonal_metrics} summarizes the seasonal-average forecasting performance. Among the evaluated models, SARIMAX exhibits the best performance.

\subsection{Temperature Effect}

This section analyzes the influence of temperature on daily waste generation using the four predictive models. The temperature values are grouped into discrete ranges to examine their relationship with average waste generation.

Figure~\ref{fig:temp_effect} illustrates the average municipal solid waste generation across twelve temperature intervals (three bins per season) ranging from 23$^\circ$C to 35$^\circ$C . The selected range covers the observed temperature range in the dataset. This subgroup-based analysis enables the variation in waste generation and forecasting performance to be examined under different temperature conditions within each season. For each temperature bin, the average actual waste generation and the average forecasted waste generation for all four models are calculated.

From Figure~\ref{fig:temp_effect}, we can see that the actual waste generation varies across different temperature ranges and therefore seasons. The observation in Figure~\ref{fig:temp_effect} is consistent with Figure~\ref{fig:seasonal_comparison}, where Summer and Winter show comparatively higher waste generation, while Monsoon and Post-Monsoon show lower levels. Thus, the seasonal and temperature-wise analyses are broadly aligned.

%

\subsection{Humidity Effect}

This section examines the influence of humidity on daily waste generation using the four predictive models.
The observed humidity conditions in the dataset are divided into eight season-wise humidity intervals, with two intervals considered for each season. Within each humidity interval, the average actual waste generation and the corresponding average forecasted waste generation are calculated. 

\begin{figure}
	\centering
	
	\begin{subfigure}{\linewidth}
		\centering
		\includegraphics[width=\linewidth]{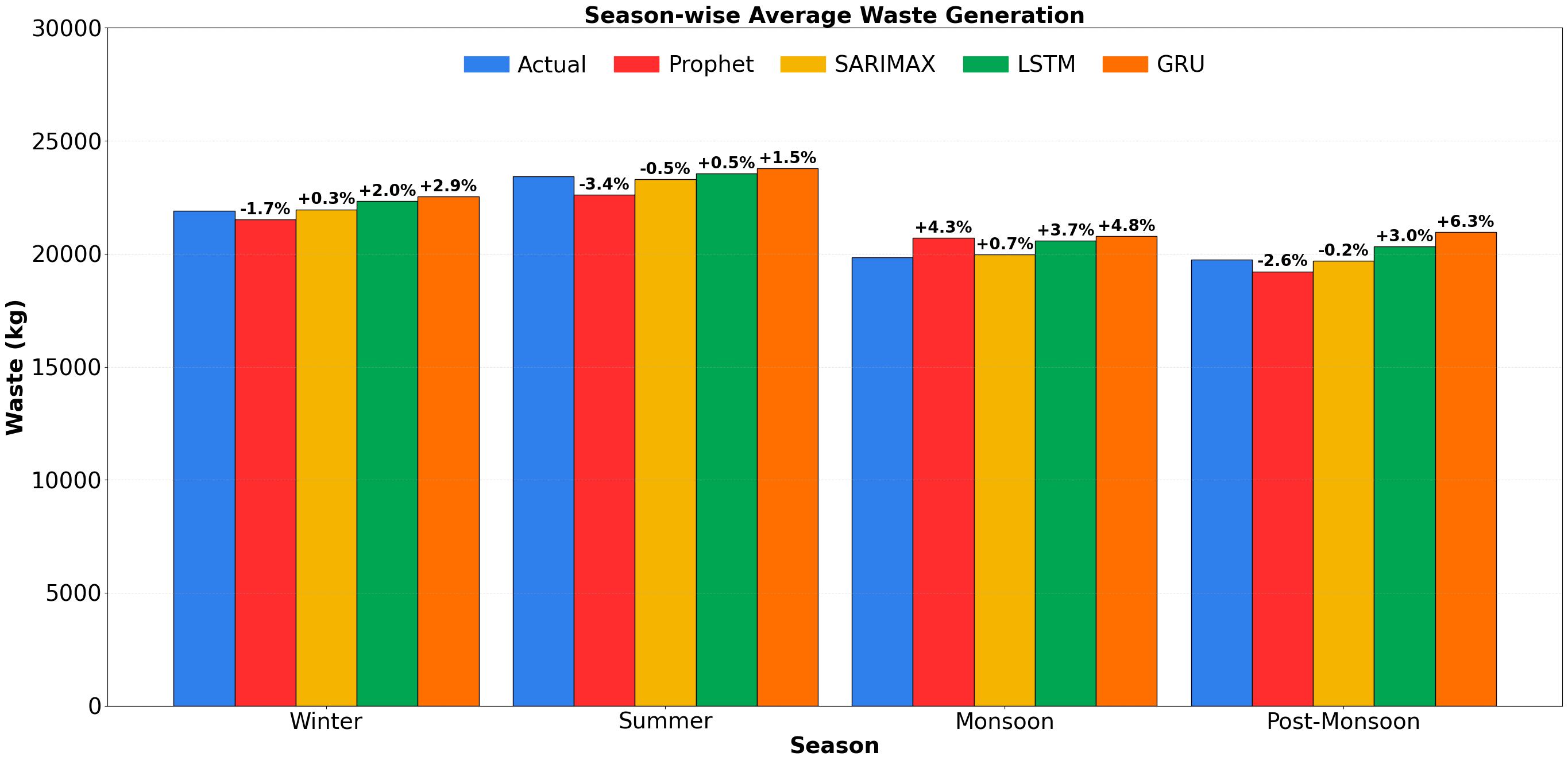}
		\caption{Season-wise comparison of actual and predicted waste generation.}
		\label{fig:seasonal_comparison}
	\end{subfigure}
	
	\vspace{0.2cm}
	
	\begin{subfigure}{\linewidth}
		\centering
		\caption{Seasonal-average forecasting performance of different models.}
		\label{fig:seasonal_metrics}
		
		\begin{tabular}{lcccc}
			\hline
			\textbf{Model} & \textbf{RMSE} & \textbf{MAE} &
			\textbf{MAPE (\%)} & \textbf{R$^2$} \\
			\hline
			Prophet & 0.0264 & 0.0252 & 3.03 & 0.8082 \\
			SARIMAX & \textbf{0.0041} & \textbf{0.0038} &
			\textbf{0.45} & \textbf{0.9955} \\
			LSTM & 0.0111 & 0.0091 & 1.15 & 0.9659 \\
			GRU & 0.0333 & 0.0301 & 3.73 & 0.6950 \\
			\hline
		\end{tabular}
	\end{subfigure}
	
	\vspace{0.3cm}
	
	\begin{subfigure}{0.48\linewidth}
		\centering
		\includegraphics[width=\linewidth]{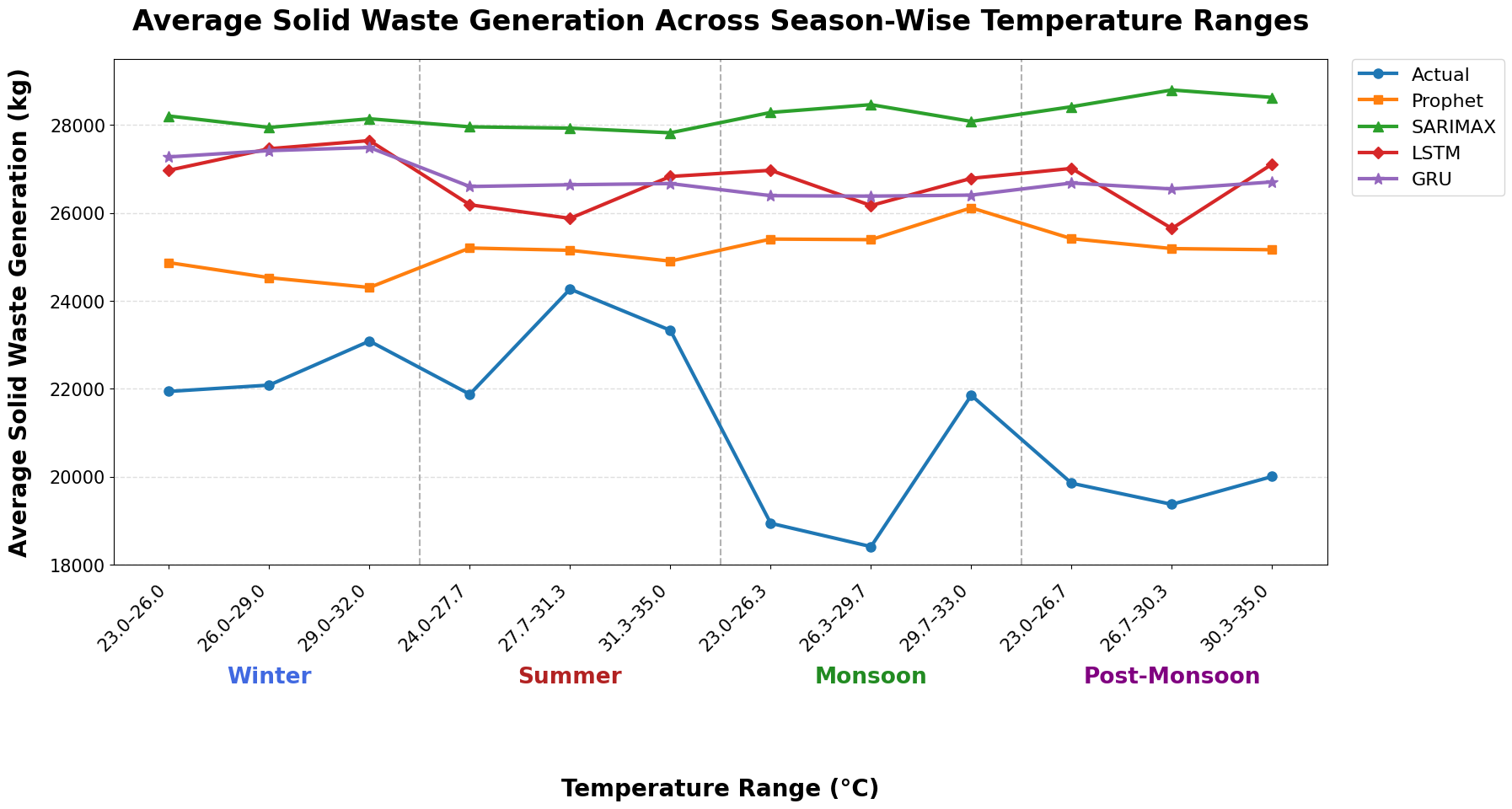}
		\caption{Temperature-wise analysis.}
		\label{fig:temp_effect}
	\end{subfigure}
	\hfill
	\begin{subfigure}{0.48\linewidth}
		\centering
		\includegraphics[width=\linewidth]{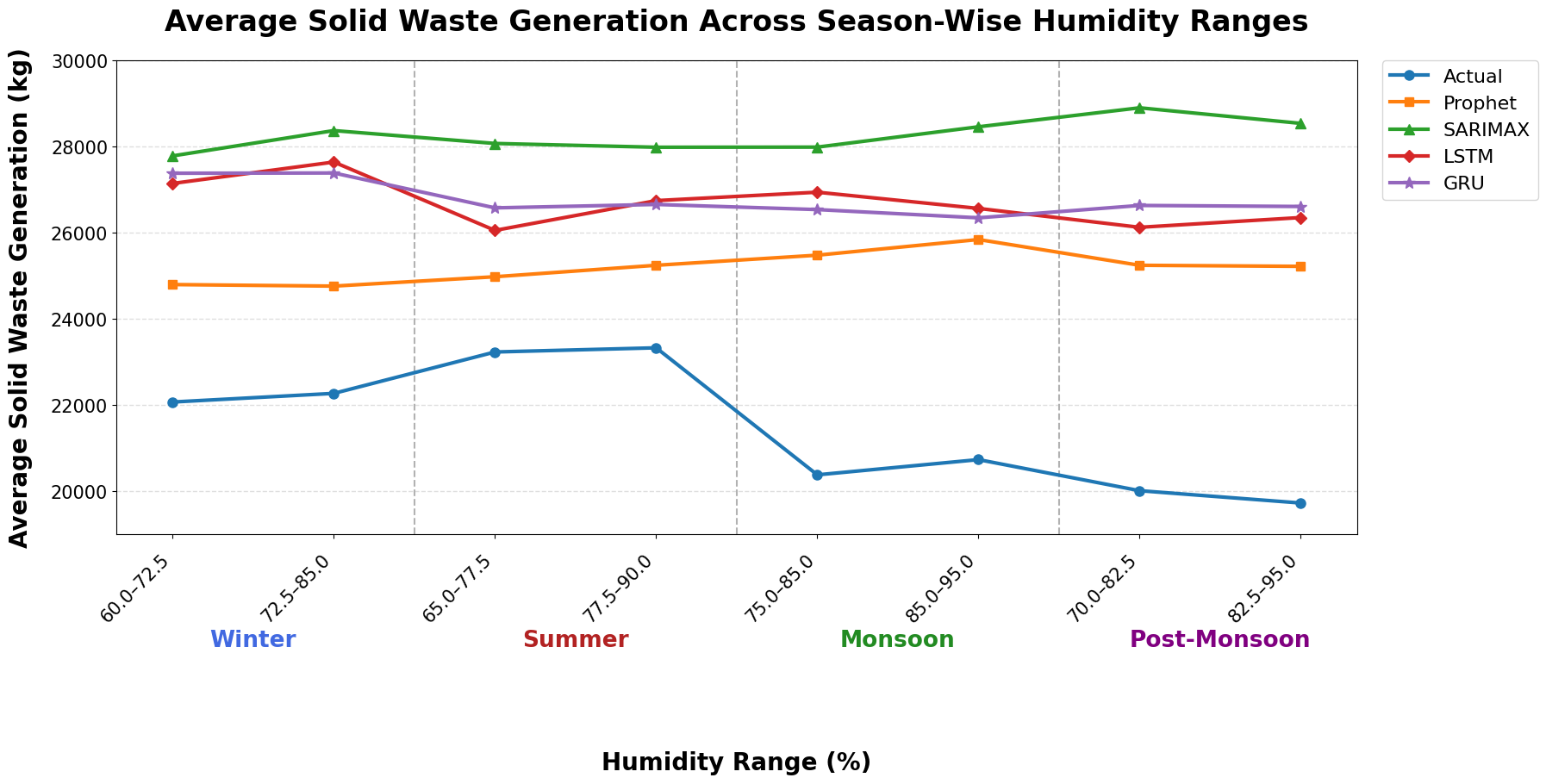}
		\caption{Humidity-wise analysis.}
		\label{fig:humidity_effect}
	\end{subfigure}
	
	\caption{Seasonal and weather(temperature and humidity) related analysis of municipal solid waste generation}
	\label{fig:combined_seasonal_weather}
\end{figure}

Figure~\ref{fig:humidity_effect} presents the average waste generation across the season-wise humidity ranges for the actual observations and the four forecasting models. The actual observations show that waste generation is comparatively higher during Winter and Summer, whereas lower values are observed during Monsoon and Post-Monsoon. The humidity-wise results are also broadly aligned with the seasonal comparison presented in Figure~\ref{fig:seasonal_comparison}. 


%
%
%

\subsection{Overall Model Forecasting}

\begin{figure}
	\centering
	
	\begin{subfigure}{0.48\linewidth}
		\centering
		\includegraphics[width=\linewidth,height=4.5cm,keepaspectratio]{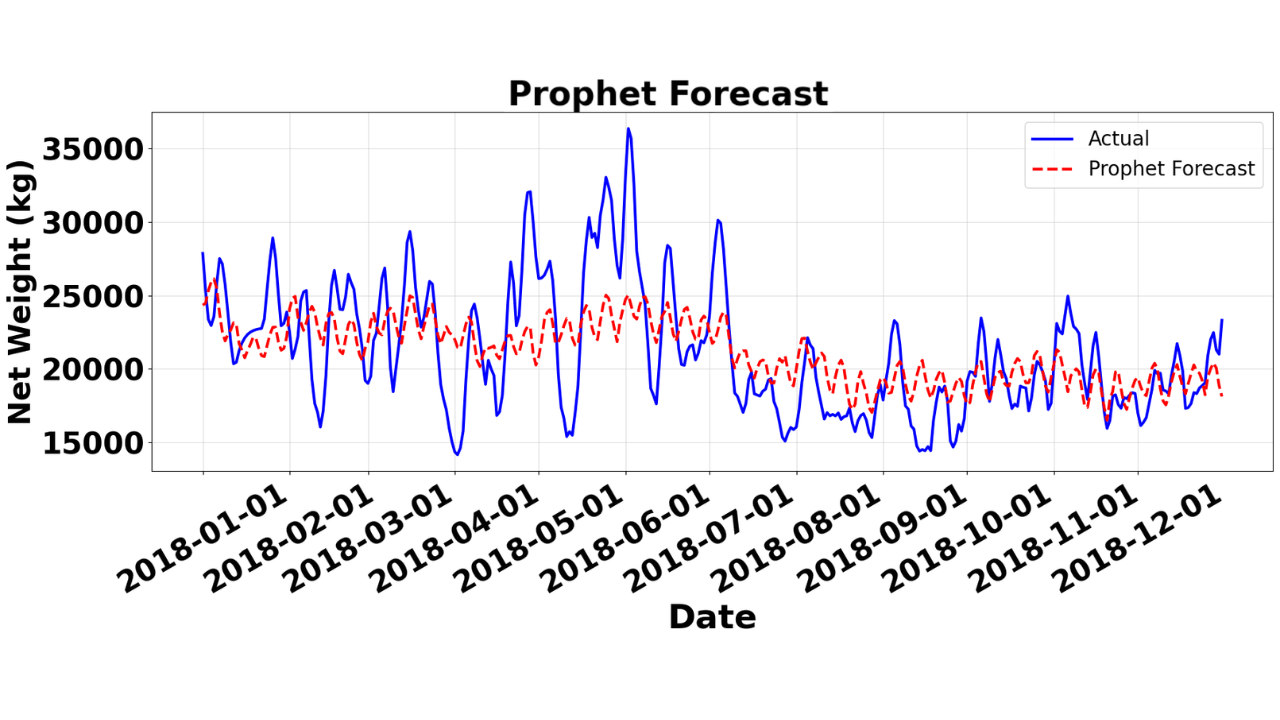}
		\caption{Prophet}
		\label{fig:prophet_forecast}
	\end{subfigure}
	\hfill
	\begin{subfigure}{0.48\linewidth}
		\centering
		\includegraphics[width=\linewidth,height=4.5cm,keepaspectratio]{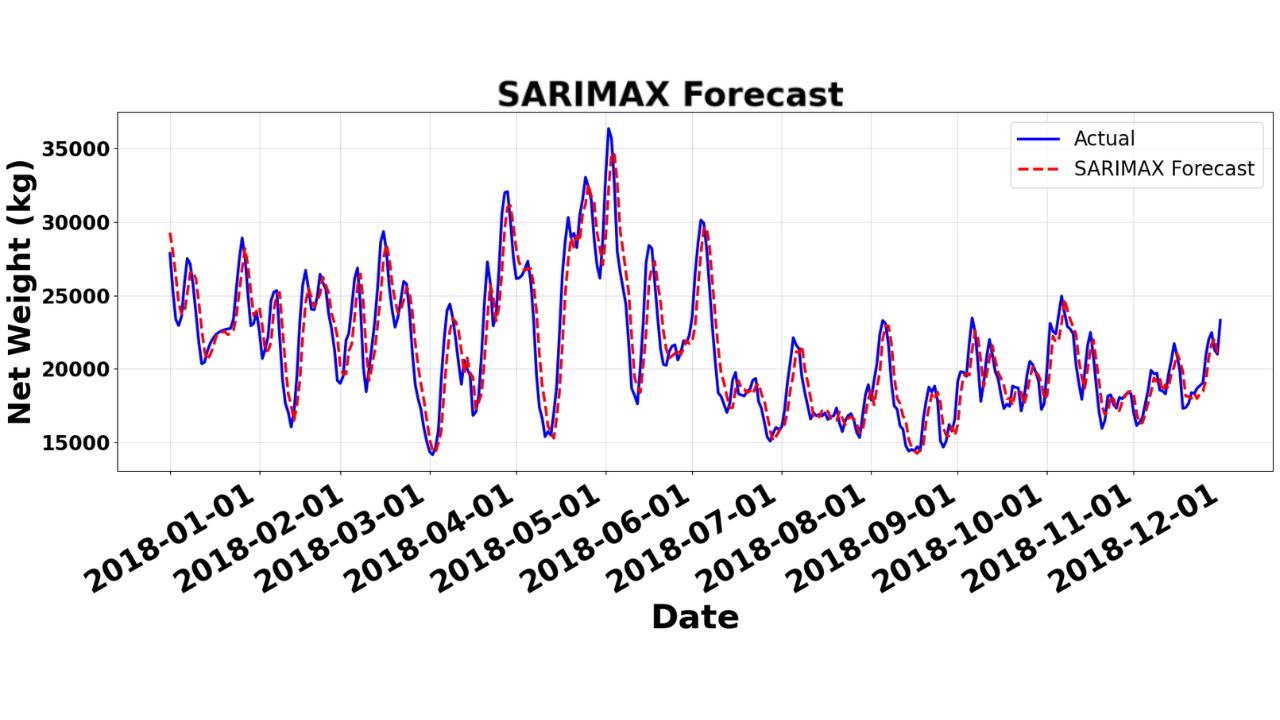}
		\caption{SARIMAX}
		\label{fig:sarimax_forecast}
	\end{subfigure}
	
	\vspace{0.3cm}
	
	\begin{subfigure}{0.48\linewidth}
		\centering
		\includegraphics[width=\linewidth,height=4.5cm,keepaspectratio]{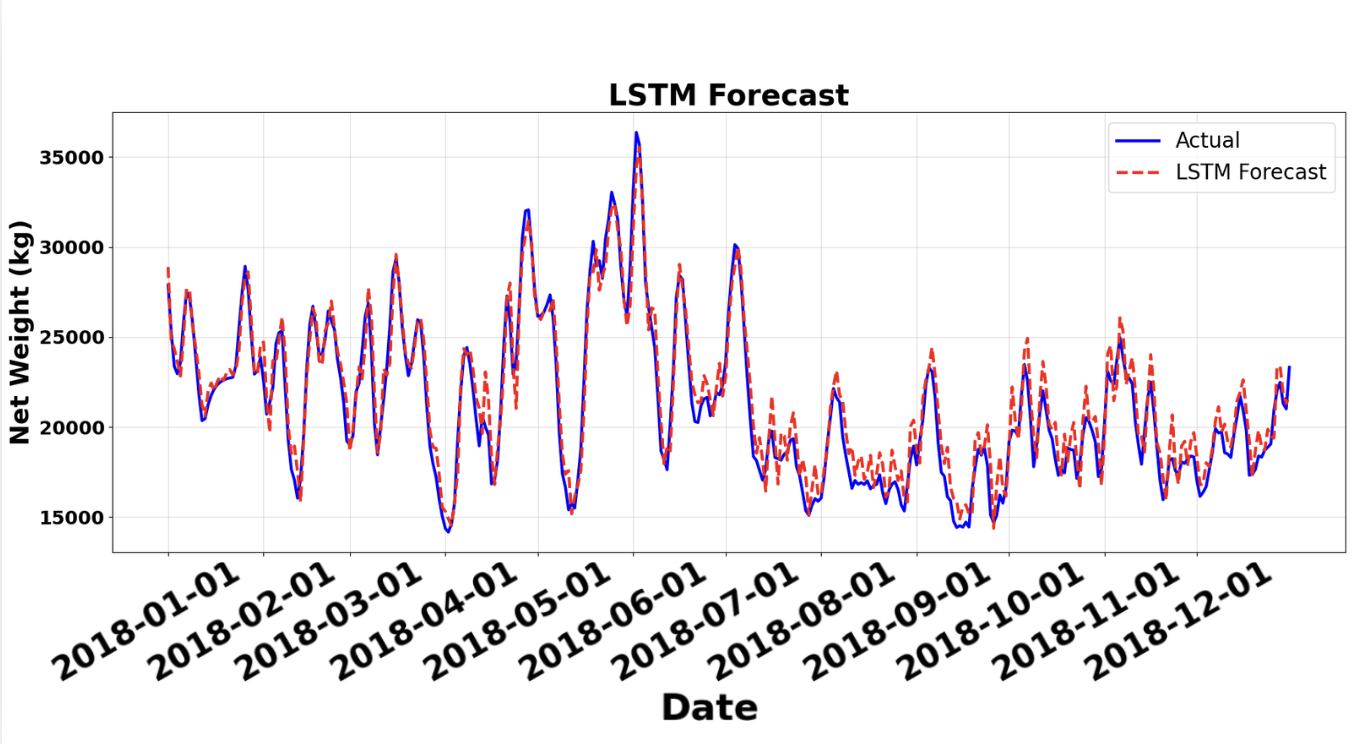}
		\caption{LSTM}
		\label{fig:LSTM_forecast}
	\end{subfigure}
	\hfill
	\begin{subfigure}{0.48\linewidth}
		\centering
		\includegraphics[width=\linewidth,height=4.5cm,keepaspectratio]{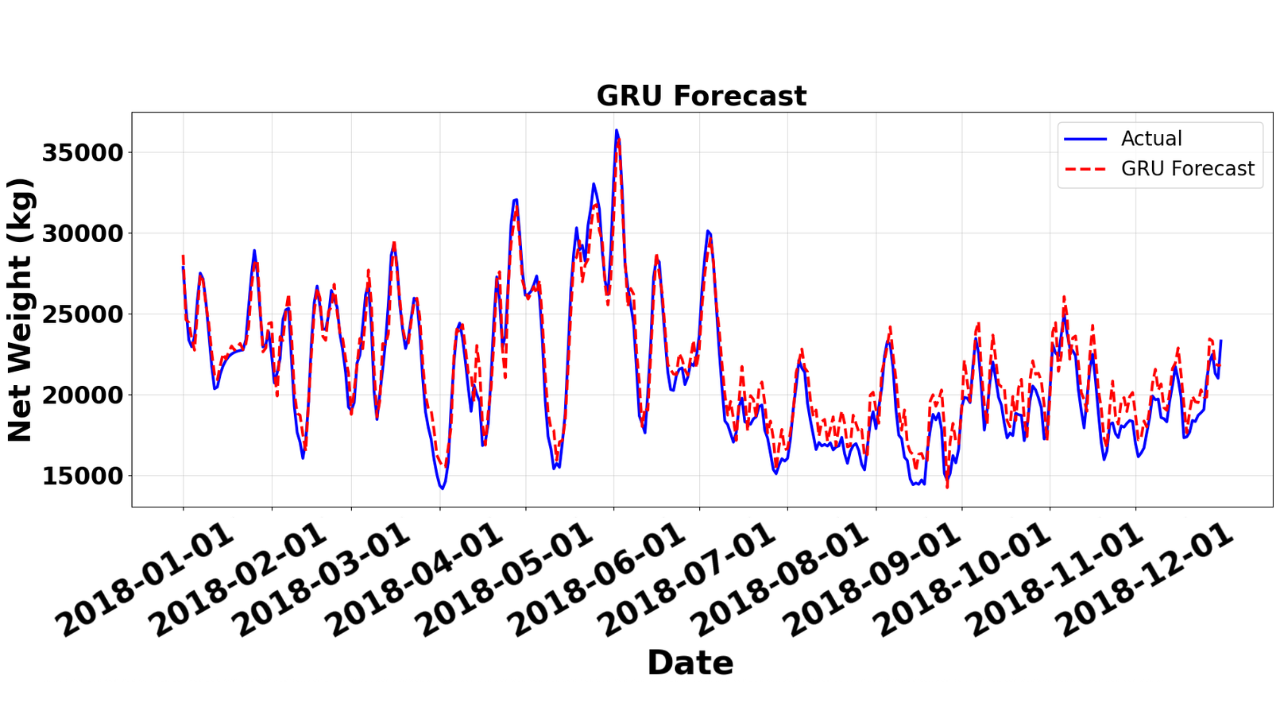}
		\caption{GRU}
		\label{fig:gru_forecast}
	\end{subfigure}
	
	\caption{Comparison of actual and forecasted municipal solid waste generation using different models }
	\label{fig:combined_forecast}
\end{figure}

Figure~\ref{fig:combined_forecast} illustrates the comparison between actual and forecasted municipal solid waste generation for 2018. LSTM, GRU and SARIMAX show closer alignment with the actual data, while Prophet exhibits poorer performance.  This observation is further supported by Table~\ref{tab:error}, where the best values are highlighted.

%

\subsection{Summarization}

This section summarizes the key findings by identifying variations in waste generation under different influencing factors and highlighting the best-performing model in each case. 
In terms of the holiday effect, waste generation decreases during holidays. For holiday periods, LSTM produces best performance to capture this effect.
For weekday and weekend patterns, waste generation is higher during weekdays. SARIMAX is the only model that preserves this weekday-dominant pattern.
Seasonal analysis reveals that waste generation is highest during Summer, followed by Winter, whereas Monsoon and Post-Monsoon exhibit comparatively lower waste generation. SARIMAX provides the closest predictions to the actual seasonal averages across all seasons.
Temperature and humidity do not show a linear relationship with waste generation. However, seasonal temperature/humidity variations are associated with changes in waste generation in different seasons. 
Table~\ref{tab:summary} summarizes the variation in waste generation across different factors along with the best-performing models in each case. 

%
%
%
%
%
%
%
%

\begin{table}
	\centering
	\caption{Overall forecasting performance and factor-wise summary of waste generation prediction.}
	\label{tab:combined_summary}
	\scriptsize
	
	\begin{subtable}[t]{0.48\linewidth}
		\centering
		\caption{Forecasting performance of different models}
		\label{tab:error}
		\begin{tabular}{lcccc}
			\hline
			\textbf{Model} & \textbf{RMSE} & \textbf{MAE} &
			\textbf{MAPE (\%)} & $\mathbf{R^2}$ \\
			\hline
			Prophet & 0.1090 & 0.0848 & 8.65 & 0.5571 \\
			SARIMAX & 0.0630 & 0.0502 & 5.05 & 0.8520 \\
			LSTM & \textbf{0.0441} & \textbf{0.0364} &
			\textbf{4.62} & \textbf{0.9329} \\
			GRU & 0.0474 & 0.0398 & 5.11 & 0.9225 \\
			\hline
		\end{tabular}
	\end{subtable}
	\hfill
	\begin{subtable}[t]{0.48\linewidth}
		\centering
		\caption{Factor-wise prediction summary}
		\label{tab:summary}
		\begin{tabular}{p{1.3cm} p{2.3cm} p{1.2cm}}
			\hline
			\textbf{Factor} & \textbf{Effect on waste generation} &
			\textbf{Best model} \\
			\hline
			Holiday & Lower on holidays & LSTM \\
			\hline
			Weekday/ \\Weekend & Lower on weekends & SARIMAX \\
			\hline
			Seasonal & Highest in Summer, followed by Winter; lower in Monsoon and Post-Monsoon & SARIMAX \\
			\hline
			Overall & -- & LSTM \\
			\hline
		\end{tabular}
	\end{subtable}
	
\end{table}

\section{Conclusion }
\label{con}
This work presents a comprehensive comparative evaluation of four widely used time-series forecasting models, namely Prophet, SARIMAX, LSTM, and GRU, using a real-world MSW dataset from Sri Lanka spanning the years 2012–2018. The dataset is enhanced by incorporating additional factors such as temperature, humidity and holiday indicators, to better capture seasonal and temporal variations in waste generation. 
Beyond quantitative evaluation, this work presents a detailed factor-wise analysis considering holidays, weekday–weekend variations, seasonal effects and environmental influences like the effect of temperature and humidity on waste generation. The results show lower waste generation during holidays and weekends, higher generation during Summer and Winter, and comparatively lower generation during Monsoon and Post-Monsoon. Temperature and humidity do not exhibit a direct linear relationship with waste generation, although their seasonal variations are associated with changes in waste generation across different seasons. The analysis provides valuable interpretability into how socio-temporal and climatic factors impact waste generation behaviour. Among all evaluated methods, the LSTM and SARIMAX models achieve the best overall performance.

Future research may therefore focus on developing hybrid forecasting frameworks that combine the complementary strengths of statistical and deep learning models to better capture complex temporal and seasonal variations in waste generation. Further studies may also incorporate additional socio-economic and demographic factors, such as household income, employment rate, household size, consumption patterns and GDP per capita, to improve the understanding of waste generation behaviour.

				
				\bibliographystyle{cas-model2-names}
				
				\bibliography{Bibliography}

				
%
%
				
			\end{document}